\documentclass[letterpaper]{article}
\usepackage{aaai}
\usepackage{times}
\usepackage{helvet}
\usepackage{courier}
\usepackage{times}
\usepackage{latexsym}
\usepackage{multirow}
\usepackage[linesnumbered]{algorithm2e}
\usepackage{amsmath}
\usepackage{amssymb}
\usepackage{graphicx}
\usepackage{tabularx}
\usepackage{natbib}
\usepackage{CJKutf8}
\usepackage[OT2,T1]{fontenc}
\usepackage{xcolor}
\input{cyracc.def}
\newcommand\textcyr[1]{{\fontencoding{OT2}\fontfamily{wncyr}\selectfont #1}}
\begin{document}
%
\title{A Unified Query-based Generative Model for \\ Question Generation and Question Answering}
\author{Linfeng Song$^1$, Zhiguo Wang$^2$ \and Wael Hamza$^2$ \\
$^1$Department of Computer Science, University of Rochester, Rochester, NY 14627 \\
$^2$IBM T.J. Watson Research Center, Yorktown Heights, NY 10598 \\
}
\maketitle
\begin{abstract}
\begin{quote}
We propose a query-based generative model for solving both tasks of question generation (QG) and question answering (QA).
The model follows the classic encoder-decoder framework. 
The encoder takes a passage and a query as input then performs query understanding by matching the query with the passage from multiple perspectives. 
The decoder is an attention-based Long Short Term Memory (LSTM) model with copy and coverage mechanisms.
In the QG task, a question is generated from the system given the passage and the target answer, whereas in the QA task, the answer is generated given the question and the passage.
During the training stage, we leverage a policy-gradient reinforcement learning algorithm to overcome exposure bias, a major problem resulted from sequence learning with cross-entropy loss.
For the QG task, our experiments show higher performances than the state-of-the-art results. 
When used as additional training data, the automatically generated questions even improve the performance of a strong extractive QA system.
In addition, our model shows better performance than the state-of-the-art baselines of the generative QA task.
\end{quote}
\end{abstract}

\section{Introduction}

Recently both question generation and question answering tasks are receiving increasing attention from both the industrial and academic communities.
The task of question generation (QG) is to generate a fluent and relevant question given a passage and a target answer, while the task of question answering (QA) is to generate a correct answer given a passage and a question.
Both tasks have massive industrial values: QA has been used in industrial products such as search engines, while QG is helpful for improving QA systems by automatically increasing the training data. It can also be used to generate questions for educational purposes such as language learning.

For the QG task, existing work either entirely ignores the target answer \citep{du2017learning} while generating the corresponding question, or directly hard-codes the answer positions into the passage \citep{zhou2017neural,yang2017semi,subramanian2017neural,tang2017question,wang2017joint,yuan2017machine}, so that sequence-to-sequence model \citep{sutskever2014sequence} can be simply utilized.
These methods only highlight the answer positions, but neglect other potential interactions between the passage and the target answer.
In addition, this kind of methods will shrivel when the target answer does not  occur in the passage verbatim.
For the QA task, most of the existing literatures \citep{wang2016machine,wang2016multi,shen2016reasonet,wang-EtAl:2017:Long2,chen-EtAl:2017:Long4,xiong2016dynamic,seo2016bidirectional,lee2016learning,yu2016end,dhingra-EtAl:2017:Long2} focus on the extractive QA scenario, where they assume that the target answer occurs in the passage verbatim. The task then is to extract a span of consecutive words from the passage as the final answer. 
However, these methods may not work well on the generative QA scenario, where the correct answer is not a span in the given passage.

\begin{figure*}[tbp]
\centering
\includegraphics[scale=0.42]{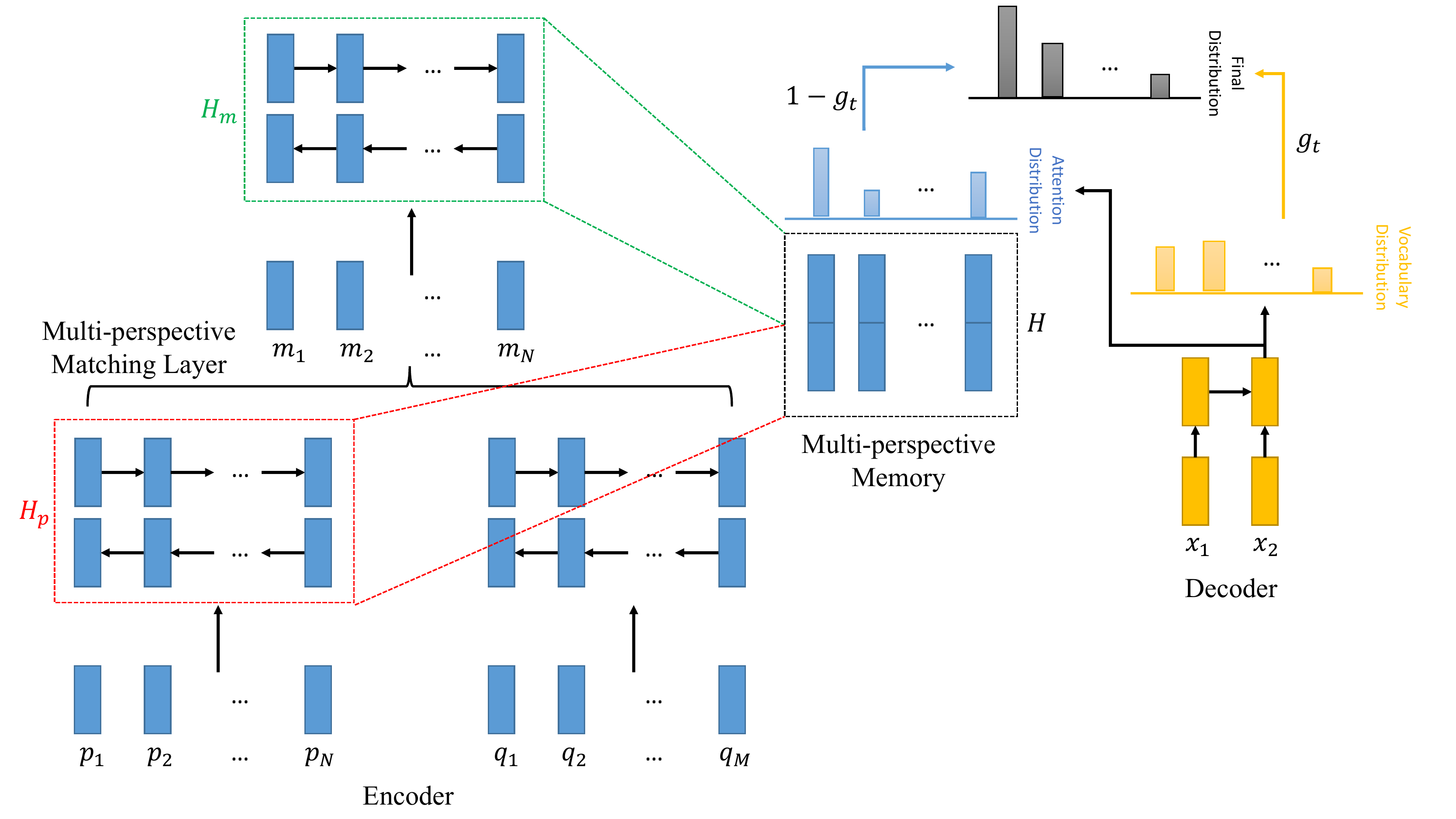}
\caption{Model overview.}
\label{fig:model}
\end{figure*}

In this paper, we cast both the QG and QA tasks into one process by firstly matching the input passage against the query, then generating the output according to the matching results.
Our model follows the classic encoder-decoder framework, where the encoder takes a passage and a query as input, then performs query understanding by matching the query with the passage from multiple perspectives, and the decoder is an attention-based LSTM model with copy \citep{gulcehre-EtAl:2016:P16-1,gu-EtAl:2016:P16-1,see2017get} and coverage \citep{tu2016modeling,mi2016coverage} mechanisms.
Here a perspective is a way of matching the query and the passage.
In the QG task, the input query is the target answer, and the decoder generates a question for the target answer, whereas in the QA task, the input query is a question, and the decoder generates the corresponding answer. 
To the best of our knowledge, there is no existing work dealing with both tasks using the same framework.
In the QG task, we are the first to investigate query understanding before generating questions. By matching the target answer against the passage from multiple perspectives, our model captures more interactions between the answer and the passage, so that it can generate more precise question for the answer. Moreover, our model does not require that the answer literally occurs in the passage.
In the QA task, our model generates answers word by word, and it has the capacity to generate answers that do not literally occur in the passage. Therefore, it naturally works for the generative QA scenario.

We first pretrain the model with the cross-entropy loss, then fine tune with policy-gradient reinforcement learning to alleviate the exposure bias problem, resulting from sequence learning with the cross-entropy loss.
In our policy-gradient reinforcement learning algorithm, we adopt a similar sampling strategy as the scheduled sampling strategy \citep{bengio2015scheduled} for generating the sampled output.
We perform experiments on the SQuAD dataset \citep{rajpurkar-EtAl:2016:EMNLP2016} for the QG task, and on the ``description'' subset of the MS-MARCO \citep{nguyen2016ms} dataset for the generative QA task.
Experimental results on the QG task show that our model outperforms previous state-of-the-art methods, and the automatically generated questions can even improve an extractive QA system.
For the generative QA task, our model shows better performance than other generative systems.

\section{Model}

Figure \ref{fig:model} shows the architecture of our model. 
The model takes two components as input: a passage $P=(p_1, ..., p_j, ..., p_N)$ of length $N$, and a query $Q=(q_1, ..., q_i, ..., q_M)$ of length $M$, then generates the output sequence 
$X=(x_1, ..., x_L)$ word by word. 
Specifically, the model follows the encoder-decoder framework. 
The encoder matches each time-step of the passage against all time-steps of the query from multiple perspectives, and encodes the matching result into a ``Multi-perspective Memory''. 
In addition, the decoder generates the output sequence one word at a time based on the ``Multi-perspective Memory''.

\subsection{Multi-Perspective Matching Encoder}
The left-hand side of Figure \ref{fig:model} depicts the architecture of our encoder. Its goal is to perform comprehensive understanding of the query and the passage. The encoder first represents all words within the passage and the query with word embeddings \citep{mikolov2013efficient}. 
In order to incorporate contextual information into the representation of each time-step of the passage or the query, we utilize a bi-directional LSTM (BiLSTM) \citep{hochreiter1997long} layer to encode the passage and the query individually:
\begin{align*}
\overleftarrow{h^q_i} &= \mathrm{LSTM}(\overleftarrow{h^q_{i+1}},q_i) \\
\overrightarrow{h^q_i} &= \mathrm{LSTM}(\overrightarrow{h^q_{i-1}},q_i) \\
\overleftarrow{h^p_j} &= \mathrm{LSTM}(\overleftarrow{h^p_{j+1}},p_j) \\
\overrightarrow{h^p_j} &= \mathrm{LSTM}(\overrightarrow{h^p_{j-1}},p_j) \textrm{,}
\end{align*}
where $q_i$ and $p_j$ are embedding of the $i$-th word in the query and the $j$-th word in the passage.
Then, the contextual vectors for each time-step of the query and the passage are constructed by concatenating the outputs from the BiLSTM layer: $h^q_i = [\overleftarrow{h^q_i};\overrightarrow{h^q_i}]$ and $h^p_j = [\overleftarrow{h^p_j};\overrightarrow{h^p_j}]$.

\begin{figure}[tbp]
\begin{center}
\includegraphics[width=0.4\textwidth]{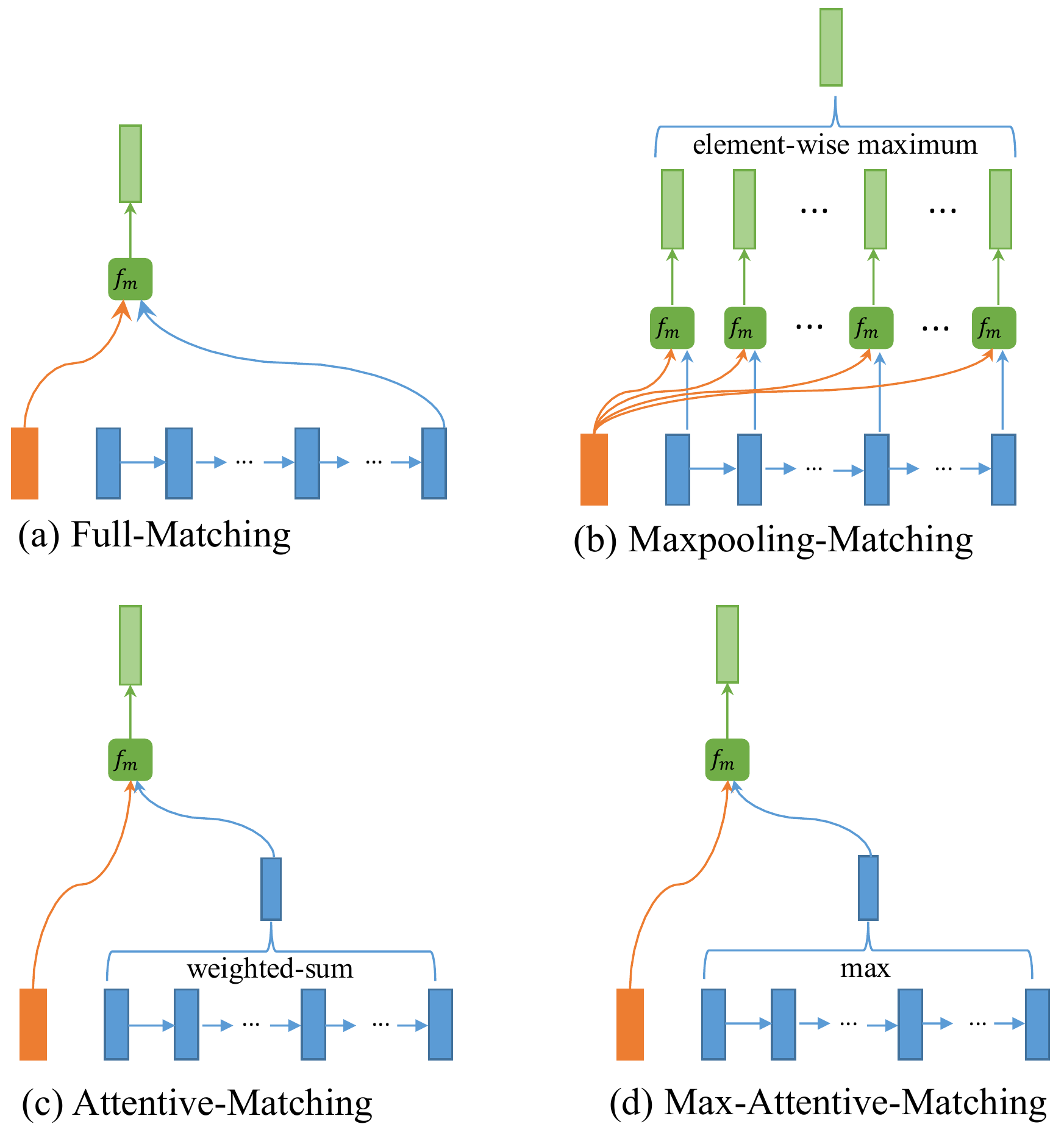}
\end{center}
\caption{Diagrams for different matching strategies, where $f_m$ is a matching function between two vectors. The inputs include the contextual vector of one time-step of the passage (left orange block) and the contextual vectors of all time-steps of the query (right blue blocks). The output is a vector of matching values (top green block) calculated via $f_m$.}
\label{fig:match_strategies}
\end{figure}

We utilize a matching layer on top of the contextual vectors to match each time-step of the passage with all time-steps of the query. 
Apparently, this is the most crucial layer in our encoder. 
Inspired by \cite{wang2017bilateral}, we adopt the multi-perspective matching method for the matching layer. 
We define four matching strategies, as shown in Figure \ref{fig:match_strategies}, to match the passage with the query from multiple granularities.  

(1) \textbf{Full-Matching}. As shown in Figure \ref{fig:match_strategies} (a), each forward (or backward) contextual vector of the passage is compared with the last time-step of the forward (or backward) representation of the query. 

(2) \textbf{Maxpooling-Matching}. As shown in Figure \ref{fig:match_strategies} (b), each forward (or backward) contextual vector of the passage is compared with every forward (or backward) contextual vectors of the query, and only the maximum value of each dimension is retained.

(3) \textbf{Attentive-Matching}. As shown in Figure \ref{fig:match_strategies} (c), we first calculate the cosine similarities between each forward (or backward) contextual vector of the passage and every forward (or backward) contextual vector of the question.  
Then, we take the cosine similarities as the weights, and calculate an attention vector for the entire query by computing a weighted sum of all the contextual vectors of the query.
Finally, we match each forward (or backward) contextual vector of the passage with its corresponding attentive vector.

(4) \textbf{Max-Attentive-Matching}. As shown in Figure \ref{fig:match_strategies} (d), this strategy is similar to the Attentive-Matching strategy. However, instead of taking the weighed sum of all the contextual vectors as the attentive vector, we pick the contextual vector with the highest cosine similarity as the attentive vector. Then, we match each contextual vector of the passage with its new attentive vector.

These four match strategies require a function $f_m$ to match two vectors. Theoretically, any functions for matching two vectors would work here. Inspired by \cite{wang2016multi}, we adopt the multi-perspective cosine matching function defined as:
\[
m = f_m(v_1, v_2; \mathbf{W}) \textrm{,}
\]
where $v_1$ and $v_2$ are $d$-dimensional input vectors, $\mathbf{W} \in \mathbb{R}^{l \times d}$ is the learnable parameter of multi-perspective weight, and $l$ is the number of perspectives.
Each row $W_k \in \mathbf{W}$ represents the weights associated with one perspective, and the similarity according to that perspective is defined as:
\[
m_k = \cos(W_k \circ v_1, W_k \circ v_2) \textrm{,}
\]
where $\circ$ is the element-wise multiplication operation.
So $f_m(v_1, v_2; \mathbf{W})$ represents the matching results between $v_1$ and $v_2$ from all perspectives.
Intuitively, each perspective calculates the cosine similarity between two input vectors, and it is associated with a weight vector trained to highlight different dimensions of the input vectors.
This can be regarded as considering different part of the semantics captured in the vector.

The final matching vector $m_j$ for each time-step of the passage is the concatenation of the matching results of all four strategies.
We also employ another BiLSTM layer on top of the matching layer to smooth the matching results.
We concatenate the contextual vectors, $h_j^p$, of the passage and matching vectors to be the \emph{Multi-perspective Memory} $\mathbf{H}$,
which contains both the passage information and the matching information.

\subsection{LSTM Decoder}

The right-hand side of Figure \ref{fig:model} is our decoder. Basically, it is an attention-based LSTM model \citep{bahdanau2014neural} with copy and coverage mechanisms. 
The decoder takes the ``Multi-perspective Memory'' as the attention memory, and generates the output one word at a time. 

Concretely, while generating the $t$-th word $x_t$, the decoder considers five factors as the input: 
(1) the ``Multi-perspective Memory'' $\mathbf{H}=\{h_0, ..., h_i, ..., h_N\}$, where each vector $h_i \in \mathbf{H}$ aligns to the $i$-th word in the passage; 
(2) the previous hidden state of the LSTM model $s_{t-1}$; 
(3) the embedding of previously generated word $x_{t-1}$; 
(4) the previous context vector $c_{t-1}$, which is calculated from the attention mechanism with $\mathbf{H}$ being the attentional memory; 
and (5) the previous coverage vector $u_{t-1}$, which is the accumulation of all attention distributions so far. When $t=0$, we initialize $s_{-1}$, $c_{-1}$ and $u_{-1}$ as zero vectors, and fix $x_{-1}$ to be the embedding of the sentence start token ``<s>''.

For each time-step $t$, the decoder first feeds the concatenation of the previous word embedding $x_{t-1}$ and context vector $c_{t-1}$ into the LSTM model to update the hidden state:
\begin{align*}
s_{t} &= \textrm{LSTM}(s_{t-1},[x_{t-1},c_{t-1}])
\end{align*}
Second, the attention distribution $\alpha_{t,i}$ for each time-step of the ``Multi-perspective Memory'' $h_i \in \mathbf{H}$ is calculated with the following equations:
\begin{align*}
e_{t,i} &= v_e^T \tanh(W_h h_i + W_s s_t + W_u u_{t-1} + b_{e}) \\
\alpha_{t,i} &= \frac{\exp(e_{t,i})}{\sum_{j=1}^N\exp(e_{t,j})} 
\end{align*}
where $W_h$, $W_s$, $W_v$, $b_e$ and $v_e$ are learnable parameters.
The coverage vector $u_t$ is then updated by  $u_t = u_{t-1} + \alpha_t$. And the new context vector $c_t$ is calculated via:
\begin{align*}
c_t = \sum_{i=1}^N \alpha_{t,i} h_{i}
\end{align*}
Then, the output probability distribution over a vocabulary of words at the current state is calculated by:
\[
P_{vocab} = \textrm{softmax}(V_2(V_1[s_t,c_t]+b_1)+b_2)\textrm{,}
\]
where $V_1$, $V_2$, $b_1$ and $b_2$ are learnable parameters. 
The number of rows in $V_2$ represents the number of words in the vocabulary.

On top of the LSTM decoder, we adopt the copy mechanism \citep{gulcehre-EtAl:2016:P16-1,gu-EtAl:2016:P16-1,see2017get} to integrate the attention distribution into the final vocabulary distribution.
The probability distribution is defined as the interpolation between two probability distributions:
\[
P_{final} = g_t P_{vocab} + (1-g_t) P_{attn}\textrm{,}
\]
where $g_t$ is the switch for controlling generating a word from the vocabulary or directly copying it from the passage. 
$P_{vocab}$ is the generating probability distribution as defined above, and 
$P_{attn}$ is calculated based on the attention distribution $\alpha_t$ by merging probabilities of duplicated words.
Intuitively, $g_t$ is relevant to the current decoder state, the attention results and the input. Therefore, inspired by \cite{see2017get}, we define it as:
\[
g_t = \sigma(w_c^T c_t + w_s^T s_t + w_x^T x_{t-1} + b_{g})\textrm{,}
\]
where vectors $w_c$, $w_s$, $w_x$ and scalar $b_{g}$ are learnable parameters.

\section{Policy Gradient Reinforcement Learning via Scheduled Sampling}
A common way of training a sequence generation model is to optimize the log-likelihood of the gold-standard output sequence $Y^*=y_0^*, ..., y_t^*, ..., y_T^*$ with the cross-entropy loss:
\[
l_{ce} = -\sum_{t=1}^T \log p(y_t^*|y_{t-1}^*,...,y_0^*,X;\theta)\textrm{,}
\]
where $X$ is the model input, and $\theta$ represents the trainable model parameters.

However, this method suffers from two main issues. First, during the training stage, the ground-truth of the previous word $y_{t-1}^*$ is taken as the input to predict the probabilities of the next word $y_t$. But, in the testing stage, the ground-truth $y_{t-1}^*$ is not available, and the model has to rely on the previously generated word $y_{t-1}$. If the model selected a different $y_{t-1}$ than the ground-truth $y_{t-1}^*$, then the following generated sequence could deviate from the gold-standard sequence. This issue is known as the ``exposure bias problem''. Second, models trained with the cross-entropy loss are optimized for the log-likelihood of a sequence which is different from the evaluation metrics.

\begin{algorithm}[t] \small
 \KwData{gold-standard sequence $Y^*$}
 \KwData{greedy search sequence $\hat{Y}$}
 \KwResult{sampled sequence $Y^s$}
 $S \leftarrow$ []\;
 \For{$i$ \textbf{\upshape{in}} \upshape{range(len($Y^*$))}}{
    \uIf{$i$ $<$ \upshape{len($\hat{Y}$)}}{
        \uIf{\upshape{random.random() $<$ $p_{flip}$}}{
        	$S$\upshape{.append($\hat{Y}$[$i$])} \;
        }\Else{
        	$S$\upshape{.append($Y^*$[$i$])} \;
        }
    }\Else{
        $S$\upshape{.append($Y^*$[$i$])} \;
    }
 }
 \caption{Scheduled Sampling Strategy}
 \label{algo:sample}
\end{algorithm}

In this work, we utilize a reinforcement learning method to address the exposure bias problem and directly optimize the evaluation metrics. Concretely, we adopt the ``REINFORCE with a baseline'' algorithm \citep{williams1992simple}, a well-known policy-gradient reinforcement learning algorithm, to train our model, because it has shown the effectiveness for several sequence generation tasks \citep{paulus2017deep,rennie2016self}. Formally, the loss function is defined as:
\[
l_{rl} = (r(\hat{Y})-r(Y^s)) \sum_{t=1}^{T} \log p(y^s_t|y^s_{t-1},...,y^s_0,x;\theta)\textrm{,}
\]
where $Y^s=y^s_0, ..., y^s_T$ is the sampled sequence, $\hat{Y}$ is the sequence generated from a baseline, and the function $r(Y)$ is the reward calculated based on the evaluation metric. 
Intuitively, the loss function $l_{rl}$ enlarges the log-probability of the sampled sequence $Y^s$, if $Y^s$ is better than the baseline $\hat{Y}$ in terms of the evaluation metric $r(Y)$, or vice versa.
In this work, for the QG task, we use the BLEU score \citep{papineni-EtAl:2002:ACL} as the reward, and for the QA task, we use the ROUGE score \citep{lin2004rouge} as the reward. 

Following \cite{rennie2016self}, we take the greedy search result from the current model as the baseline sequence $\hat{Y}$.
\cite{rennie2016self} generated the sampled sequence $Y^s$ according to the probability distribution of $p(y^s_t|y^s_{t-1},...,y^s_1,x;\theta)$. 
However, this sampling strategy doesn't work well for our tasks. 
One possible reason is that our tasks have much larger search space. 
Inspired by \cite{bengio2015scheduled}, we designed a new ``Scheduled Sampling'' strategy to construct the sampled sequence $Y^s$ from both the gold-standard sequence $Y^*$ and the greedy search sequence $\hat{Y}$.
As shown in Algorithm \ref{algo:sample}, it goes through the gold-standard sequence $Y^*$ word by word (Line 2), and replaces with the corresponding word from the greedy search sequence $\hat{Y}$ with probability $p_{flip}$ (Line 4-8).
If the greedy search sequence is shorter than the gold-standard sequence, the ground-truth word is used after exceeding the end of the greedy search sequence (Line 10).
Our experiments show that sampling the sequence according to the model distribution, as \cite{rennie2016self} does, usually produces outputs worse than the greedy search sequence, so it does not help very much.
On the other hand, our sampling strategy usually generates better outputs than the greedy search sequence.

\begin{table*}
\centering
\begin{tabular}{l|ccc|c}
\hline
\multirow{2}{*}{Models} & \multicolumn{3}{c|}{Split 1} & Split 2 \\
                        & BLEU-4 & METEOR & ROUGE-L  & BLEU-4 \\
\hline
\cite{du2017learning}    & 12.28 & 16.62 & 39.75 & --    \\
\cite{zhou2017neural}    & --    & --    & --    & 13.29 \\
~~~~w/o rich feature (baseline) & --    & --    & --    & 12.59(*) \\
\hline
MPQG            & 12.84 & 18.02 & 41.39 & 13.39 \\
MPQG+R          & \textbf{13.98} & \textbf{18.77} & \textbf{42.72} & \textbf{13.91} \\
\hline
\end{tabular}
\caption{Results on question generation. *There is no published scores for \cite{zhou2017neural} without the rich features, so we re-implemented their system and show the result.}
\label{tab:qg_main}
\end{table*}

\section{Experimental Setup}

We conduct experiments on two tasks: question generation (QG) and generative question answering (QA).

\textbf{Question Generation}
For the QG task, we evaluate the quality of generated questions with some automatic evaluation metrics such as BLEU \citep{papineni-EtAl:2002:ACL} and ROUGE \citep{lin2004rouge}, as well as their effectiveness in improving an extractive QA system.
We conduct experiments on the SQuAD dataset \citep{rajpurkar-EtAl:2016:EMNLP2016} by comparing our model with \cite{du2017learning} and \cite{zhou2017neural} in terms of BLEU, METEOR \citep{banerjee2005meteor} and ROUGE.
The dataset contains 536 articles and over 100k questions related to the articles.
Here, we follow \cite{du2017learning} and \cite{zhou2017neural} to conduct experiments on the accessible part as our entire dataset.
Since \cite{du2017learning} and \cite{zhou2017neural} conducted their experiments using different training/dev/test split, we conduct experiments on both splits, and compare with their reported performance.

In addition, we also evaluate our model from a more practical aspect by examining whether the automatically generated questions are helpful for improving an extractive QA system.
We use the data split of \cite{du2017learning}, and conduct experiments on low-resource settings, where only (10\%, 20\%, or 50\%) of the human-labeled questions in the training data are available.
For example, in the 10\% setting, we first train our QG model with the 10\% available training data, then generate questions for the remaining 90\% instances in the training data, where the human-labeled questions are abandoned.\footnote{We assume the gold answers are available when generating questions for the remaining 90\% instances, and leave automatic answer selection as future work, since the primary goal here is to evaluate the quality of automatically generated questions.}
Finally, we train an extractive QA system with the 10\% human-labeled questions and the 90\% automatically generated questions. 
The extractive QA system we choose is \cite{wang2016multi}, but our framework does not make any assumptions about the extractive QA systems being used.

\textbf{Generative QA}
For this task, we conduct experiments on the MS MARCO dataset \citep{nguyen2016ms}, which contains around 100k queries and 1M passages.
The design purpose of this dataset is to generate the answer given the top 10 returned documents from a search engine, where the answer is not necessary in the documents. 
Even though the answers in this dataset are human generated rather than extracted from candidate documents, we found that the answers of around 66\% questions can be exactly matched in the passage, and a large number of the remaining answers just have a small difference with the content in the passages.\footnote{They are generated by dropping or paraphrasing one span from the supporting sentence (containing answer) in the passage.}
Among all types of questions (``numeric'', ``entity'', ``location'', ``person'' and ``description''), the ``description'' subset has the most percentage of answers that can not be exactly matched in the passage.
Therefore, for the generative QA experiments, we follow \cite{nguyen2016ms} to conduct experiments on the ``description'' subset, and compare with their reported results.

For both tasks, our model is first trained for 15 epochs with the cross-entropy loss, then fine-tuned for 15 epochs using our policy gradient algorithm.
Adam \citep{kingma2014adam} is used for parameter optimization, and the learning rate is set to $0.005$ and $0.0001$ for cross entropy and policy gradient phases respectively.
The encoder and decoder share the same pre-trained word embeddings, which are the 300-dimensional GloVe \citep{pennington-socher-manning:2014:EMNLP2014} 
word vectors pre-trained from the 840B common crawl corpus, and the embeddings are not updated during training.
For all experiments, the flip probability $p_{flip}$ is set to 0.1, 
the number of perspectives $l$ is set to 5, 
and the weight for the coverage loss $\eta$ is set to 0.1.
For all experiments, the model yielding the best performance on the dev set is picked for evaluation on the test set.

\section{Experimental Results}

\begin{table}[t!] \small
\centering
\begin{tabularx}{0.49\textwidth}{X} 
\hline
\textbf{Passage:} nikola tesla -lrb- serbian cyrillic : \textcyr{Nikola Tesla} ; 10 july [\emph{1856}] -- 7 january 1943 -rrb- was a serbian american inventor , electrical engineer , mechanical engineer , physicist , and futurist best known for his contributions to the design of the modern alternating current -lrb- ac -rrb- electricity supply system .\\
\textbf{Target Answer:} 1856\\
\textbf{Reference:} when was nikola tesla born ?\\
\textbf{Baseline:} when was nikola tesla 's inventor ?\\
\textbf{MPQG:} when was nikola tesla born ?\\
\textbf{MPQG+R:} when was nikola tesla born ?\\
\hline
\textbf{Passage:} zh\'{e}ng -lrb- chinese : \begin{CJK*}{UTF8}{gbsn}正\end{CJK*} -rrb- meaning `` [\emph{right}] '' , `` just '' , or `` true '' , would have received the mongolian adjectival modifiers , creating `` jenggis '' , which in medieval romanization would be written `` genghis '' .\\
\textbf{Target Answer:} right\\
\textbf{Reference:} what does zh\'{e}ng mean ? \\
\textbf{Baseline:} what are the names of the `` jenggis '' ? \\
\textbf{MPQG:} what does zh\'{e}ng UNK mean ? \\
\textbf{MPQG+R:} what does zh\'{e}ng mean ? \\
\hline
\textbf{Passage:} kenya is known for its [\emph{safaris , diverse climate and geography , and expansive wildlife reserves}] and national parks such as the east and west tsavo national park , the maasai mara , lake nakuru national park , and aberdares national park .\\
\textbf{Target Answer:} safaris , diverse climate and geography , and expansive wildlife reserves\\
\textbf{Reference:} what is kenya known for ? \\
\textbf{Baseline:} what are the two major rivers that are known for the east and west tsavo national park ? \\
\textbf{MPQG:} what is kenya known for ? \\
\textbf{MPQG+R:} what is kenya known for ? \\
\hline
\end{tabularx}
\caption{\small{Examples of generated questions. In each passage, the target answer is italic and is within brackets. The baseline is our implementation of \cite{zhou2017neural} without rich features.}}
\label{tab:qg_example}
\end{table}

\begin{table*}
\centering
\begin{tabular}{lccc|ccc}
\hline
\multirow{2}{*}{Methods} & \multicolumn{3}{c}{F1} & \multicolumn{3}{c}{Exact Match (EM)} \\ \cline{2-7}
                         & 10\% & 20\% & 50\% & 10\% & 20\% & 50\% \\
\hline
baseline  & 61.61 & 68.38 & 73.67 & 50.54 & 57.63 & 64.13 \\
w/ window & 61.23 & 66.80 & 73.35 & 50.48 & 56.31 & 64.00 \\
w/ MPQG+R & \textbf{64.52} & \textbf{69.28} & \textbf{74.50} & \textbf{55.44} & \textbf{59.66} & \textbf{65.30} \\
\hline
\end{tabular}
\caption{Results on improving extractive QA with automatically generated questions.}
\label{tab:for_extract}
\end{table*}

\subsection{Question Generation}
We compare our model with \cite{du2017learning} and \cite{zhou2017neural} on the question generation task, and show the results in Table \ref{tab:qg_main}.
Since \cite{zhou2017neural} adopts rich features (such as named entity tags and part-of-speech tags), we re-implement a version without these rich features (w/o rich feature) for fair comparison.
We also implement two versions of our model: (1) MPQG is our model only trained with the cross-entropy loss, and (2) MPQG+R is our model fine-tuned with the policy gradient reinforcement learning algorithm after pretraining.

First, our MPQG model outperforms the comparing systems on both data splits, which shows the effectiveness of our multi-perspective matching encoder. 
Our MPQG model, which only takes word features, shows even better performance than the feature-rich (with POS and NE tags) system of \cite{zhou2017neural}.
\cite{du2017learning} utilized the sequence-to-sequence \citep{peng2016recurrent} model to take the passage as input and then generated the questions, where they entirely ignored the target answer. 
Therefore, the generated questions are independent of the target answer. 
\cite{zhou2017neural} hard-coded the target answer positions into the passage, and employed the sequence-to-sequence model to consume the position-encoded passages, then generated the questions. 
This method only considered the target answer positions, but neglected the relations between the target answer and other parts of the passage. 
If the target answer does not literally occur in the passage, this method will shrivel. Conversely, our MPQG model matches the target answer against the passage from multiple perspectives. Therefore, it can capture more interactions between the target answer and the passage, and result in a more suitable question for the target answer.

Second, our MPQG+R model works better than the MPQG model on both splits, showing the effectiveness of our policy gradient training algorithm.

To better illustrate the advantage of our model, we show some comparative results of different models in Table \ref{tab:qg_example}, where the \emph{Baseline} system is our implementation of \cite{zhou2017neural} without rich features.
Generally, our MPQG model generates better questions than \cite{zhou2017neural}.
Taking the first case as an example, the baseline fails to recognize that ``1856'' is the year when ``nikola tesla'' is born,
while our MPQG learns that from the pattern ``day month year - day month year'', which frequently occurs in the training data.
For the third case, the baseline fails to generate the correct output as the query is very long and complicated.
On the other hand, our MPQG model is able to capture that, because it performs comprehensive matching between the target answer and the passages.
In addition, our MPQG+R model fixes some small mistakes of MPQG by directly optimizing the evaluation metrics, such as the second case in Table \ref{tab:qg_example}.

\subsection{Question Generation for Extractive QA}

Table \ref{tab:for_extract} shows the results on improving an extractive QA system with automatically generated questions.
Here F1 measures the overlap between the prediction and the reference in terms of bags of tokens, 
and exact match (EM) measures the percentage where the prediction is identical to the reference \citep{rajpurkar-EtAl:2016:EMNLP2016}.
The baseline is trained only on the part where gold questions are available, while the others are trained on the combination of the gold questions and the automatically generated questions, but with different methods of generating questions:
(1) \emph{w/ window}, a strong baseline from \cite{yang2017semi}, uses the previous and the following 5 tokens of the target answer as the pseudo question, and (2) \emph{w/ MPQG+R} generates questions with our MPQG+R model.

First, we can see that \emph{w/ MPQG+R} outperforms the baseline under all settings in terms of both F1 and EM scores, especially under the 10\% setting, where we observe 3 and 5 points gains in terms of F1 and EM scores.
This shows the effectiveness of our model.
Second, the comparing results between  \emph{w/ MPQG+R} and \emph{w/ window} show that the improvements of \emph{w/ MPQG+R} are not due to simply enlarging the training data, but because of the higher quality of the generated questions.
\cite{yang2017semi} showed that \emph{w/ window} can significantly improve their baseline, while it is not true in our experiment.
One reason could be that our baseline is much stronger than theirs.
For example, our system achieves 50.54\% EM score under 10\% setting, while theirs only got an EM score of 24.92\%.

\subsection{Generative QA}

\begin{table} 
\centering
\begin{tabular}{lc}
\hline
Models              & ROUGE-L \\
\hline
Best Passage         & 35.1  \\
Passage Ranking      & 17.7  \\
Sequence to Sequence & 8.9   \\
Memory Network       & 11.9  \\
\hline
vanilla-cosine       & 19.9  \\
MPQG                 & 31.5  \\
MPQG+R               & 32.9  \\
\hline
\end{tabular}
\caption{\small{Results on the ``description'' subset of MS-MARCO.}}
\label{tab:for_gqa}
\end{table}

For the generative QA experiment, we compare our model with the generative models in \cite{nguyen2016ms} on the ``description'' subset of MS-MARCO dataset. 
Table \ref{tab:for_gqa} shows the corresponding performance.
Among all the comparing methods, \emph{Best Passage} selects the best passage in terms of the ROUGE-L score, and obviously it accesses the reference.
\emph{Passage Ranking} ranks the passage by a deep structured semantic model of \cite{huang2013learning}. 
\emph{Sequence to Sequence} is a vanilla sequence-to-sequence model \citep{sutskever2014sequence}.
\emph{Memory Network} adopts the end-to-end memory network \citep{sukhbaatar2015end} as the encoder, and a vanilla RNN model as the decoder.
We also implement a baseline system ``vanilla-cosine'', which only apply the vanilla cosine similarity for the matching function $f_m$ in our encoder, and is only trained with the cross-entropy loss. 

First, we can see that our \emph{MPQG+R} model outperforms all other systems by a large margin, and is close to \emph{Best Passage}, even though \emph{Best Passage} accesses the reference.
Besides, our \emph{MPQG} model outperforms the \emph{vanilla-cosine} model showing the effectiveness of our multi-perspective matching encoder.
Finally, \emph{MPQG+R} outperforms \emph{MPQG} by around 1.4 ROUGE-L points, showing the effectiveness of our policy-gradient learning strategy.

\section{Related Work}

For question generation (QG), our work extends previous work \citep{du2017learning,zhou2017neural,yang2017semi,subramanian2017neural,tang2017question,wang2017joint,yuan2017machine} by performing query understanding.
\cite{tang2017question,yang2017semi} joins the QG task with the QA task, but they still conduct the QG task.
The only difference is that they directly optimize the QA performance rather than a general metric (such as BLEU).
On the other hand, our model can conduct both tasks of QG and QA.

For question answering (QA), most previous works \citep{wang2016machine,wang2016multi,shen2016reasonet,wang-EtAl:2017:Long2,chen-EtAl:2017:Long4,xiong2016dynamic,seo2016bidirectional,lee2016learning,yu2016end,dhingra-EtAl:2017:Long2} focus on the extractive QA scenario, which predicts a continuous span in the passage as the answer.
Obviously, they rely on the assumption that the answer can be exactly matched in the passage. 
On the other hand, our model performs generative QA, which generates the answer word-by-word, and does not rely on this assumption.
The generative QA is valuable for studying, as we can not guarantee the assumption being true for all scenarios. 
\cite{tan2017s} claims to perform generative QA, but it still relies on an extractive QA system by generating answers from the extractive results. 
One notable exclusion is \cite{yin2015neural}, which generate factoid answers from a knowledge base (KB).
One significant difference is that their method matches the query against a KB, whereas ours performs matching against unstructured texts.
Besides, we leverage policy gradient learning to alleviate the exposure bias problem, which they also suffer from.

\section{Conclusion}

In this paper, we introduced a query-based generative model, which can be used on both question generation and question answering.
Following the encoder-decoder framework, a multi-perspective matching encoder is designed to perform query and passage understanding, and an LSTM model with coverage and copy mechanisms is leveraged as the decoder to generate the target sequence.
In addition, we leverage a policy gradient learning algorithm to alleviate the exposure bias problem, which generative models suffer from when training with the cross-entropy loss.
Experiments on both question generation and question answering tasks show superior performances of our model, which outperforms the state-of-the-art models.
From the results we conclude that query understanding is important for question generation, and that policy gradient is effective on tackling the exposure bias problem resulted by sequence learning with cross-entropy loss.
For the future work, we will consider adding adversarial data, which has been shown successful on plenty of areas\citep{peng2018jointly}.

\bibliography{aaai}
\bibliographystyle{aaai}
\end{document}